\documentclass[letterpaper, 10 pt, conference]{ieeeconf}

\IEEEoverridecommandlockouts                              
\usepackage{graphics} 

\usepackage{stmaryrd} 
\usepackage{epsfig} 
\usepackage{times} 
\usepackage{amsmath} 
\usepackage{amssymb}  
\usepackage{float}
\usepackage{seqsplit}
\usepackage{caption}
\usepackage{subcaption}
\usepackage[export]{adjustbox}
\usepackage{physics}
\usepackage{tikz}
\usepackage{balance}
\usepackage{tabu}
\usepackage{multirow,booktabs}
\usepackage[amssymb]{SIunits}
\usepackage{cite}
\usepackage[colorlinks=true,allcolors=blue]{hyperref}

\newcommand{\sref}[1]{Section \ref{#1}}
\newcommand{\fref}[1]{Fig. \ref{#1}}
\newcommand{\tref}[1]{Table \ref{#1}}

\title{\LARGE \bf
Online Visual Place Recognition via Saliency Re-identification}

\author{Han Wang, Chen Wang, and Lihua Xie
\thanks{
The work is supported by Delta-NTU Corporate Laboratory for Cyber-Physical Systems under the National Research Foundation Corporate Lab @ University Scheme.
}
\thanks{Han Wang and Lihua Xie are with the School of Electrical and Electronic Engineering,
Nanyang Technological University, 50 Nanyang Avenue, Singapore 639798.
        {\tt\small e-mail: \{wang.han,elhxie\}@ntu.edu.sg}}%
\thanks{Chen Wang is with the Robotics Institute, Carnegie Mellon University, Pittsburgh, PA 15213-3890 USA. {\tt\small e-mail: chenwang@dr.com}}
}

\begin{document}

\maketitle

\begin{abstract}
As an essential component of visual simultaneous localization and mapping (SLAM), place recognition is crucial for robot navigation and autonomous driving. Existing methods often formulate visual place recognition as feature matching, which is computationally expensive for many robotic applications with limited computing power, \textit{e.g.}, autonomous driving and cleaning robot. Inspired by the fact that human beings always recognize a place by remembering salient regions or landmarks that are more attractive or interesting than others, we formulate visual place recognition as saliency re-identification. In the meanwhile, we propose to perform both saliency detection and re-identification in frequency domain, in which all operations become element-wise. 
The experiments show that our proposed method achieves competitive accuracy and much higher speed than the state-of-the-art feature-based methods. The proposed method is open-sourced and available at \url{https://github.com/wh200720041/SRLCD.git}. 
\end{abstract}
\section{Introduction}

Visual place recognition, also known as loop closure detection, is the task to identify repetitive places or landmarks \cite{lowry2015visual,wang2020intensity} during autonomous navigation. It is of great importance for generating drift-free maps in simultaneous localization and mapping (SLAM). 
In robot navigation, trajectory estimation often comes with drifts due to the sensor imperfection and environmental variation \cite{nguyen2020loosely}. Without the capability of loop closure, estimated robot pose inevitably deviates from its true position. This subsequently leads to an unreliable localization and also increases the computational expense due to repetitive registration of landmarks \cite{lowe2004distinctive}. Therefore, to enable a drift-free localization, a well-structured SLAM system usually requires visual place recognition to associate current pose with historical frames.  

Computational cost is one of the key bottlenecks in visual place recognition \cite{arroyo2015towards,cieslewski2017efficient}. For example, in a SLAM system, the query times of incoming data grows as more places are visited and registered into the database (or map), which can subsequently slow down the system or cause memory overflow in the long run. 
Most of existing works down-scale an image (or a place) into lower dimensional features to improve computational efficiency. A popular strategy is to re-formulate the visual place recognition problem as the retrieval of the local hand-crafted features such as ORB \cite{rublee2011orb} and BRISK \cite{leutenegger2011brisk}. It has been widely used in many applications due to its high accuracy, \textit{e.g.}, brute-force (BF) matching on local descriptors 
has achieved satisfactory results on public dataset \cite{schlegel2018hbst}. However, to guarantee a high accuracy, feature-based approaches require the extraction and matching of thousands of descriptors for each image, which is computationally expensive, \textit{e.g.}, iBoW-LCD \cite{garcia2018ibow} and FABMAP \cite{cummins2008fab}. 
Moreover, discretization of local descriptors requires complicated  offline  training that is troublesome in practice.

\begin{figure}[!t]
\begin{center}  
    
    \vspace{8pt}
    \includegraphics[width=1\linewidth]{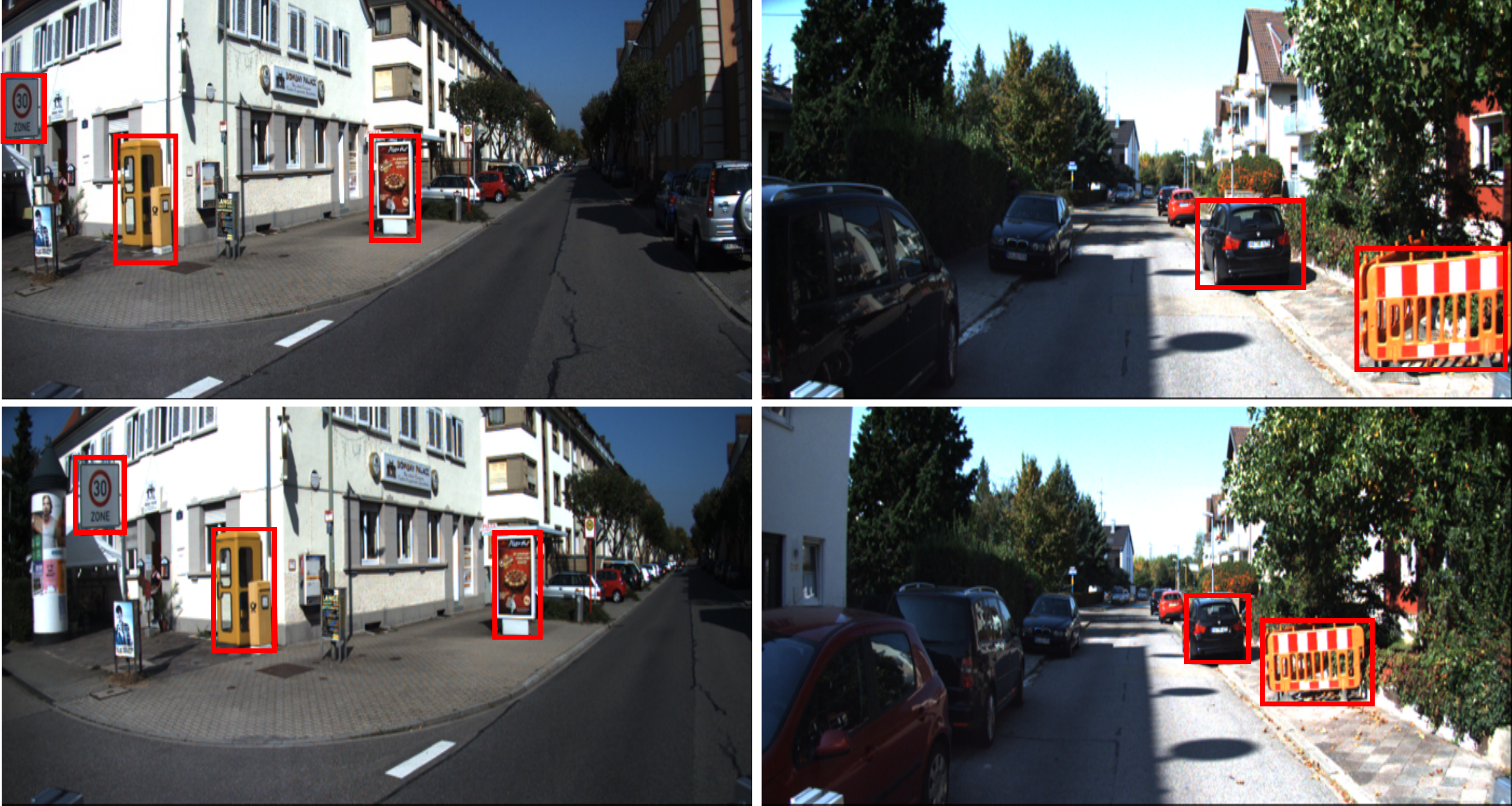}
\end{center}   
\caption{Instead of matching feature points, we formulate place recognition as saliency re-identification. Both saliency detection and retrieval are performed in frequency domain to take advantage of efficiency of element-wise operation. Different from general object detection, saliency detection extracts areas that are visually appealing. This figure shows that the salient regions are re-identified when the places are re-visited (samples from KITTI dataset \cite{geiger2013vision}).
}
\label{fig: title_graph}
\end{figure}

In many real-time robotic systems such as unmanned aerial vehicles (UAVs) and micro-robots, computational capability is very limited due to power supply and payload constraints. 
Hence an effective place representation strategy is necessary.




Inspired by the fact that human beings always recognize a place by remembering objects or landmarks that are more appealing or interesting than others, not simply by remembering point features \cite{cong2018review}, we formulate place recognition as saliency re-identification, which is more natural and straightforward than feature matching. 
However, the definition of saliency is subjective and can be varied according to one's psychological process \cite{donk2008effects} and environmental changes \cite{wang2020visual}.  
For simplicity, we follow the observation that human eyes are more sensitive to salient regions in an image with specific spectrum properties \cite{hou2007saliency}.
The salient regions can be either objects or visual distinct areas that stand out and enclosed by bounding boxes.
In this paper, we find that both saliency detection and retrieval can be performed in frequency domain, which can further reduce the computational cost.

In summary, the main contributions of this paper are:
\begin{itemize}
\item We propose a novel framework for visual place recognition for SLAM via saliency re-identification, which is coded in C++ and open sourced.
\item We propose to detect and identify the salient regions in frequency domain by taking advantage of computationally efficient element-wise operation.
\item The experiments show that our method is much faster than existing feature-based methods without reducing recall rate and precision. It does not require offline training or loading trained vocabulary.
\end{itemize}

This paper is organized as follows: \sref{sec:related-work} reviews the related works on visual place recognition. \sref{sec:methodology} describes the details of the proposed framework. \sref{sec:experiment} shows experimental results and comparison with existing methods, followed by conclusion in \sref{sec:conclusion}.
\section{Related Work}
\label{sec:related-work}

In this section, we review the most widely used feature-based methods, including both offline training-based approach and online training-based approach. We will also review the most recent deep learning-based methods.

\subsection{Feature-based Methods}
Although proposed early, feature matching is still the most popular strategy for visual place recognition, especially in robotics. 
One of the most classic approaches is the fast appearance-based mapping (FABMAP) \cite{glover2012openfabmap}.
In FABMAP, the visual vocabulary of SURF feature \cite{bay2006surf} is trained by hierarchical k-means clustering to discretize its high dimensional representation.
Revisited places are then identified by matching feature distribution over the offline trained visual vocabulary.
Similar idea is employed in DBoW2 \cite{galvez2012bags}, in which the binary descriptor ORB \cite{rublee2011orb} is used instead. It achieves faster speed due to the binary feature representation. 
However, discretization of local descriptors requires complicated offline training for visual vocabulary that is troublesome in practice. For example, in FABMAP, it takes a few hours to train such visual vocabulary on a desktop computer. Moreover, the training data cannot contain any loop and have to be collected in the similar environment in order to achieve a good result. 

Instead of building an offline trained visual vocabulary to discretize local feature space, some recent works introduce more efficient feature retrival to make online training possible. 
For example, Emilio \textit{et al.} \cite{garcia2018ibow} introduce a vocabulary maintenance strategy called dynamic island that groups similar images. The dynamic island identifies repetitive registration of same descriptor across multiple frames. The database is kept at small scale by removing those redundant descriptors.
Schlegel \textit{et al.} \cite{schlegel2018hbst} introduce an online trained Hamming distance embedded binary search tree (HBST) to image retrieval which is much faster than traditional FLANN matching methods. 
However, the memory cost is huge since raw descriptors are used to build the incremental visual vocabulary tree. 
Applying dimensional reduction on local features is also an alternative solution.
Carrasco \textit{et al.} \cite{carrasco2016global} adopt the idea from image encoding to extract local descriptors. Hash coding is applied to the local features array collected from single image so that the extracted features are more compact. 
Gehrig \textit{et al.} \cite{gehrig2017visual} directly apply principle component analysis (PCA) for BRISK \cite{leutenegger2011brisk} feature. Instead of searching on pre-trained visual vocabulary, k-nearest neighbor (K-NN) search is performed on the projected descriptors that achieves faster speed at the level of millisecond for each query. However, those methods require local feature extraction, which is computationally expensive.


\subsection{Deep Learning-based Methods}

The recent success of convolutional neural network (CNN) in computer vision \cite{wang2019kervolutional} has triggered another research trend for visual place recognition.
For example, in \cite{chen2017deep}, a multi-scale feature encoding method is introduced by training two CNN architectures. The generated CNN features are viewpoint invariant, hence a large performance improvement is achieved.
Inspired by the traditional image retrival method, \textit{i.e.}, vector of locally aggregated descriptors (VLAD), Relja \textit{et al.} propose the NetVLAD \cite{arandjelovic2016netvlad} to learn CNN parameters in an end-to-end manner for place recognition.
Hausler \textit{et al.} \cite{hausler2019multi} combine both feature-based methods and the CNN techniques. 
The query image is trained by combining sum of absolute difference (SAD), histogram of oriented gradients (HOG) \cite{dalal2005histograms}, CNN spatial Max pooling and CNN spatial Arg-Max pooling.
However, those CNN-based methods achieve high accuracy at the expense of a huge computational burden, which is sometimes infeasible for real-time systems.

\section{Methodology}
\label{sec:methodology}

The proposed framework mainly consists of two components: saliency detection and saliency retrieval, both of which will be processed in frequency domain to take advantage of the efficiency of element-wise operation.

\subsection{Saliency Detection}
Existing methods on saliency analysis mainly focus on spectral analysis \cite{cong2018review}. In particular, we introduce log-spectral residual approach to identify visual saliency which has been widely applied in practice \cite{hou2007saliency}. 
Log-spectral analysis can be computed in an efficient element-wise manner, which can subsequently reduce the computational cost.

\subsubsection{Definition}
We present the basic procedure of log-spectral residual for saliency detection. Denote the 2-D FFT $\mathcal{F}: \mathbb{C}^{m\times n} \mapsto \mathbb{C}^{m\times n}$ as $\hat{\cdot}$, given an input image $\mathbf{I}$, its FFT can be expressed as $\hat{\mathbf{I}}$. We denote the magnitude spectrum and phase spectrum as $\mathcal{A}(\hat{\mathbf{I}})$ and $\mathcal{P}(\hat{\mathbf{I}})$, respectively, thus 
the log-spectral residual $\mathbf{\mathcal{R}}(\hat{\mathbf{I}})$ can be defined as:
\begin{subequations}
\begin{align}
\mathbf{\mathcal{L}}(\hat{\mathbf{I}}) &= \log(\mathcal{A}(\hat{\mathbf{I}})), \\
\mathbf{\mathcal{R}}(\hat{\mathbf{I}}) &= \exp( \mathbf{\mathcal{L}}(\hat{\mathbf{I}}) - \mathbf{f}_{ave} * \mathbf{\mathcal{L}}(\hat{\mathbf{I}})), 
\label{Equation:saliecyresult}
\end{align}
\end{subequations}
where $\mathbf{f}_{ave}$ is a normalized average filter of size $k\times k$ 
and $*$ denotes the cross-correlation.
A saliency map $\mathbf{\mathcal{M}}$ can be simply extracted by converting the residual back to spatial domain via inverse fast Fourier transform (IFFT):
\begin{equation}\label{eq:saliency-map}
    \mathbf{\mathcal{M}} = \mathcal{G}_{\sigma} * \mathcal{F}^{-1}\left(\mathcal{R}(\hat{\mathbf{I}})\cdot \exp(\mathrm{\mathbf{j}}\cdot\mathcal{P}(\hat{\mathbf{I}}))\right),\\
\end{equation}
where $\mathcal{G}_{\sigma}$ is a Gaussian filter and $\mathrm{\mathbf{j}}$ is the imaginary unit.
This saliency map \eqref{eq:saliency-map} implies the saliency distribution of the original image \cite{srivastava2003advances}. With such distribution, we can derive saliency regions $\mathbf{x}_1, \mathbf{x}_2, \cdots  \in\mathbf{I}$ by simply taking pixel connectivity analysis on the saliency map $\mathbf{\mathcal{M}}$.

\begin{figure}[!t]
	\begin{center}
		\includegraphics[width=1.0\linewidth]{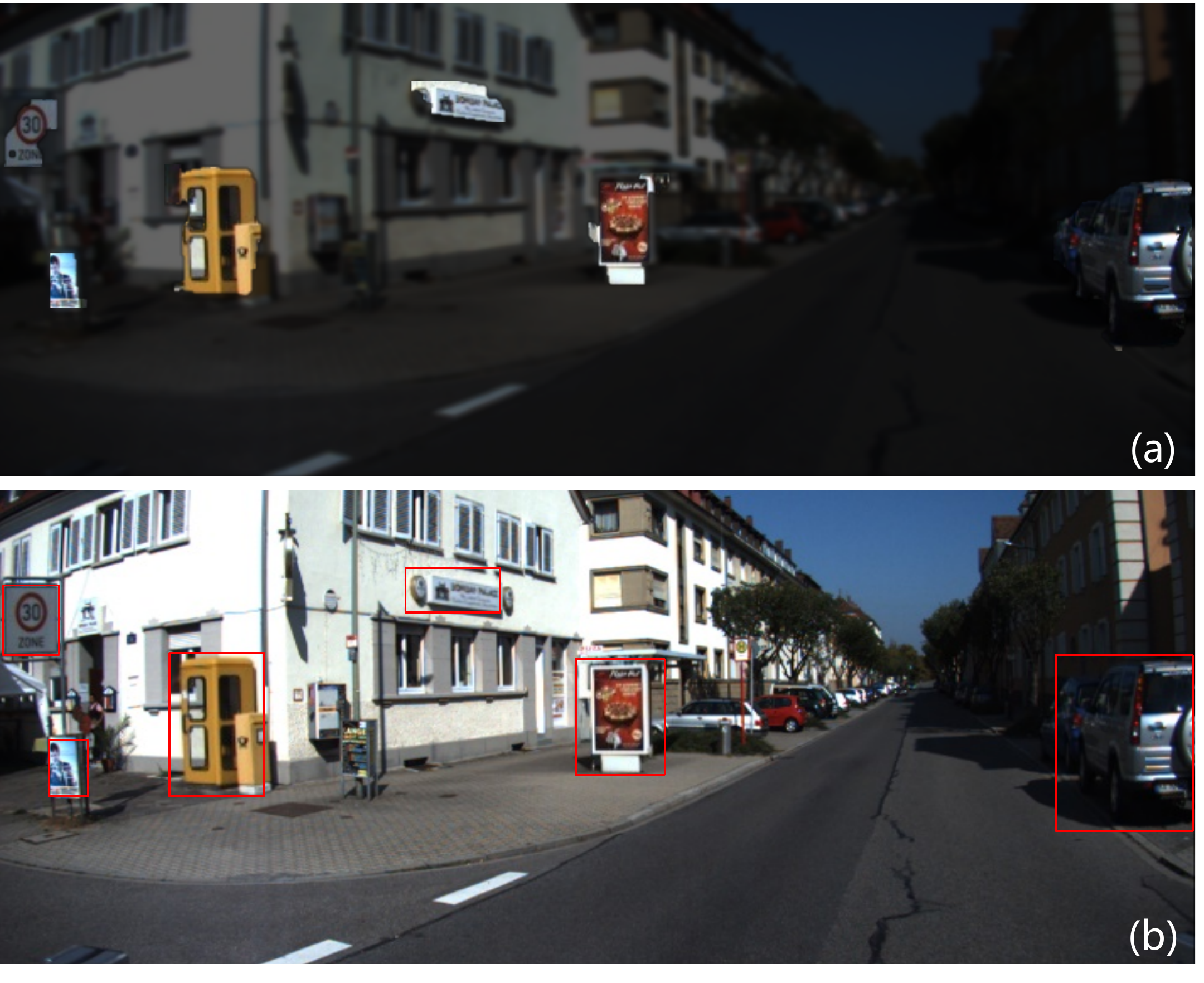}
	\end{center}
	\caption{An example of saliency detection from KITTI dataset. (a) shows the mask of the detected saliency map. (b) is the detected salient regions enclosed by red bounding boxes.}
	\label{fig: Examples of Object Extraction}
\end{figure}

The log-spectral residual approach provides a fast solution for saliency detection. However, not all extracted saliency information is suitable for place recognition, \textit{e.g.}, the regions that are too small or too dark or bright. Therefore, we need extra rules to filter out the unqualified regions in order to improve the robustness.

It is observed that the detected saliency with high contrast and more edges is often unique and highly distinguishable, which is suitable for place recognition.
Considering the computational cost, we adopt a discriminative strategy using contrast density $\phi_{x}$ and edge complexity $\rho_{x}$ defined in \eqref{eq:contrast-density}.
\begin{subequations}\label{eq:contrast-density}
\begin{align}
    & \phi_{\mathbf{x}} = \overline{(\mathbf{x} - \overline{\mathbf{x}})^2},
    \\
    & \rho_{\mathbf{x}} = \frac{\mathcal{C}(\mathbf{x})}{\mathcal{S}(\mathbf{x})} ,
\end{align}
\end{subequations}
where $\mathbf{x} \in \mathbf{I}$ is the extracted salient region, $\Bar{\mathbf{x}}$ is the mean pixel intensity, $\mathcal{C}$ is Canny edge filter, and $\mathcal{S}$ computers the total size of salient region (i.e., pixel number).
In experiment, we find that this strategy can remove most of less informative regions and even perform better than more complicated rules.

\begin{figure*}[!t]
\begin{center}
    \includegraphics[width=0.95\linewidth]{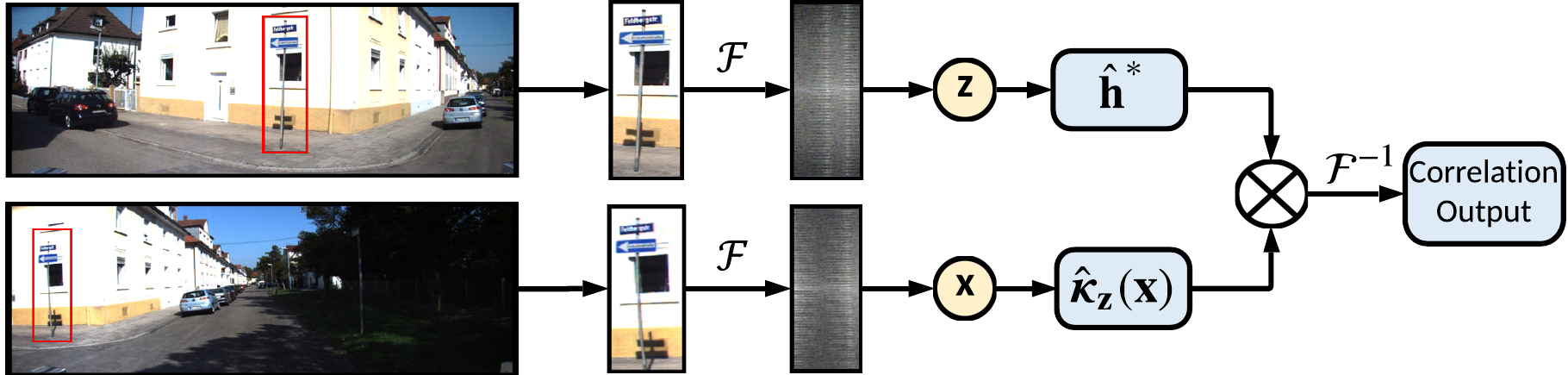}
\end{center}
\caption{The procedure of saliency retrieval. The same traffic sign is identified when a car re-visits the same place.}
\label{fig: Loop Detected}
\end{figure*}

The output of log-spectrum residual algorithm is a mask with roughly object shapes.
For better matching, we enclose those masks by minimum bounding boxes.
An example of saliency detection is shown in \fref{fig: Examples of Object Extraction}. It can be seen that six distinct regions, including traffic signs, ad boards, phone booths, and cars are extracted.
Although it is inevitable that some dynamic objects, \textit{e.g.}, cars, are extracted, they will be further removed by consistency verification in \sref{sec:geometric}.

\subsubsection{Property}
One of the reasons that we take the log-spectral residual for saliency detection is that the detected results are translation, rotation, and scale equivariant, which is crucial for visual place recognition.
Recall that the corresponding FFT of translation-rotated image $\mathbf{I}_2(x,y)=\mathbf{I}_1(x\cos\theta+y\sin\theta-x_0, -x\sin\theta +y\cos\theta-y_0)$ is related by 
$\hat{\mathbf{I}}_2(u,v)= e^{-j2\pi(ux_0+vy_0)}\cdot \hat{\mathbf{I}}_1(u\cos\theta+v\sin\theta,-u\sin\theta+v\cos\theta)$ \cite{reddy1996fft}, in which the factor $e^{-j2\pi(ux_0+vy_0)}$ does not change the magnitude spectrum. This means that log-spectrum residual $\mathbf{\mathcal{R}}(\hat{\mathbf{I}}_2)$ is just a rotation replica of $\mathbf{\mathcal{R}}(\hat{\mathbf{I}}_1)$, resulting in a rotated saliency map $\mathcal{M}$ and unchanged salient regions. This also holds for scale transforms. Suppose $\mathbf{I}_2$ is a scale transform of $\mathbf{I}_1$ with scale factor $s$, thus $\hat{\mathbf{I}}_2(u,v) = \frac{1}{s^2}\hat{\mathbf{I}}_1(u/s, v/s)$ and $\mathbf{\mathcal{L}}(\hat{\mathbf{I}}_2(u,v))=\mathbf{\mathcal{L}}(\hat{\mathbf{I}}_1(u/s, v/s))-2\log s$, meaning that the log-spectrum is scaled and the same salient region can be extracted.
Those properties are crucial for robust saliency retrieval.

\subsection{Saliency Retrieval}

Although the saliency extraction is robust and consistent, the extracted salient regions can slightly differ from size, viewing angles, illumination, \textit{etc}. 
Therefore, a transform-invariant encoding algorithm is necessary for saliency retrieval.
Considering the computational cost, we introduce the kernel cross-correlator (KCC) \cite{wang2018kernel}, which is able to match two image areas directly in frequency domain and is invariant to affine transforms such as translation, rotation, and scale. 
More specifically, we compare the current salient region $\mathbf{z}$ with the salient region $\mathbf{x}$ from database. The purpose of training is to find an optimal correlator that is unique for each salient region. Then the correlator is used to examine the similarity of two salient regions in the retrieval stage.

\subsubsection{Definition}
Recall that the cross-correlation of two 2-D signals $\mathbf{g} = \mathbf{x} * \mathbf{h}$ becomes $\hat{\mathbf{g}}= \hat{\mathbf{x}}\varodot \hat{\mathbf{h}}^*$ in frequency domain, where the operator $*$, $\varodot$ and $\mathbf{h}^*$ represent cross-correlation, element-wise multiplication, and complex conjugate, respectively. Given correlation output $\mathbf{g}$, the kernel cross-correlation is defined as:
\begin{equation}\label{eq:kcc}
 \hat{\mathbf{g}} = \hat{\boldsymbol{\kappa}}_{\mathbf{z}}(\mathbf{x}) \varodot  \hat{\mathbf{h}}^*,
\end{equation}
where $\boldsymbol{\kappa}_\mathbf{z}(\mathbf{x})$ is a kernel matrix given by:
\begin{equation}
\boldsymbol{\kappa}_\mathbf{z}(\mathbf{x})=\begin{bmatrix}
\kappa(\mathbf{x},\mathbf{z}_{11}) &\dots &\kappa(\mathbf{x},\mathbf{z}_{1n})\\
\vdots & \ddots & \vdots\\
\kappa(\mathbf{x},\mathbf{z}_{m1}) &\dots &\kappa(\mathbf{x},\mathbf{z}_{mn})
\end{bmatrix},
\end{equation}
with $\kappa(\cdot, \cdot)$ being a kernel function. $\mathbf{x}$ and $\mathbf{z}$ are the candidate salient region and target salient region. $\mathbf{z}_{ij}$ is the translational shift of $\mathbf{z}$ with $i$ pixel shifts in horizontal and $j$ pixel shifts in vertical direction. For a salient region $\mathbf{z}$ of size $m\times n$, there are $m\times n$ translational shifts in total.

Specifically, we use the Gaussian kernel $\kappa(\mathbf{x}, \mathbf{z})=\phi(\|\mathbf{x}-\mathbf{z}\|^2)$. Let function $\boldsymbol{\Phi}(\cdot)$ be the matrix form of Gaussian kernel function, where each element is a single Gaussian kernel $\phi(\cdot)$, then the kernel matrix $\boldsymbol{\kappa}_\mathbf{z}(\mathbf{x})$ becomes:
\begin{equation}\label{eq:kernel-matrix-n2}
\begin{aligned}
    \boldsymbol{\kappa}_\mathbf{z}(\mathbf{x})
    &=\begin{bmatrix}
   \boldsymbol{\phi}(\|\mathbf{x}-\mathbf{z}_{11}\|^2) &\dots &\boldsymbol{\phi}(\|\mathbf{x}-\mathbf{z}_{1n}\|^2)\\
    \vdots & \ddots & \vdots\\
    \boldsymbol{\phi}(\|\mathbf{x}-\mathbf{z}_{m1}\|^2) &\dots & \boldsymbol{\phi}(\|\mathbf{x}-\mathbf{z}_{mn}\|^2)
    \end{bmatrix}
    \\
    & = \boldsymbol{\Phi}(\|\mathbf{x}\|^2 + \|\mathbf{z}\|^2 - 2\cdot [\text{Tr}(\mathbf{x}^T\mathbf{z}_{ij})]_{mn}),
\end{aligned}
\end{equation}
where $\text{Tr}(\mathbf{\cdot})$ is the matrix trace and $[\text{Tr}(\mathbf{x}^T\mathbf{z}_{ij})]_{mn}$ denotes:
\begin{equation}
 [\text{Tr}(\mathbf{x}^T\mathbf{z}_{ij})]_{mn}=\begin{bmatrix}
   \text{Tr}(\mathbf{x}^T\mathbf{z}_{11}) &\dots &\text{Tr}(\mathbf{x}^T\mathbf{z}_{1n})\\
   \vdots & \ddots & \vdots\\
   \text{Tr}(\mathbf{x}^T\mathbf{z}_{m1}) &\dots &\text{Tr}(\mathbf{x}^T\mathbf{z}_{mn})
   \end{bmatrix},
\end{equation}
From the 2-D correlation theory, we know that $\mathbf{x} * \mathbf{z} = [\text{Tr}(\mathbf{x}^T\mathbf{z}_{ij})]_{mn}$, so that we have
\begin{subequations}               
\begin{align} \label{equation:kcc-training-a}
    \boldsymbol{\kappa}_\mathbf{z}(\mathbf{x})
                & = \boldsymbol{\Phi}(\|\mathbf{x}\|^2 + \|\mathbf{z}\|^2 - 2\cdot \mathbf{x}*\mathbf{z})\\
                & =\boldsymbol{\Phi}(\frac{\|\hat{\mathbf{x}}\|^2}{m\cdot n} + \frac{\|\hat{\mathbf{z}}\|^2}{m\cdot n} - 2\cdot \mathcal{F}^{-1}(\hat{\mathbf{x}}\varodot\hat{\mathbf{z}}^*)).\label{equation:kcc-training-c}
\end{align}
\end{subequations}
This reduces the complexity $\mathcal{O}(N^2)$ of \eqref{eq:kernel-matrix-n2}  to $\mathcal{O}(N\log N)$ of \eqref{equation:kcc-training-c}, where $N=m\times n$ is the number of pixels in the salient region. This implies that KCC can also be calculated efficiently due to the element-wise operation in frequency domain \cite{wang2018correlation}.

\begin{figure*}[!t]
	\begin{center}
		\includegraphics[width=1.0\linewidth]{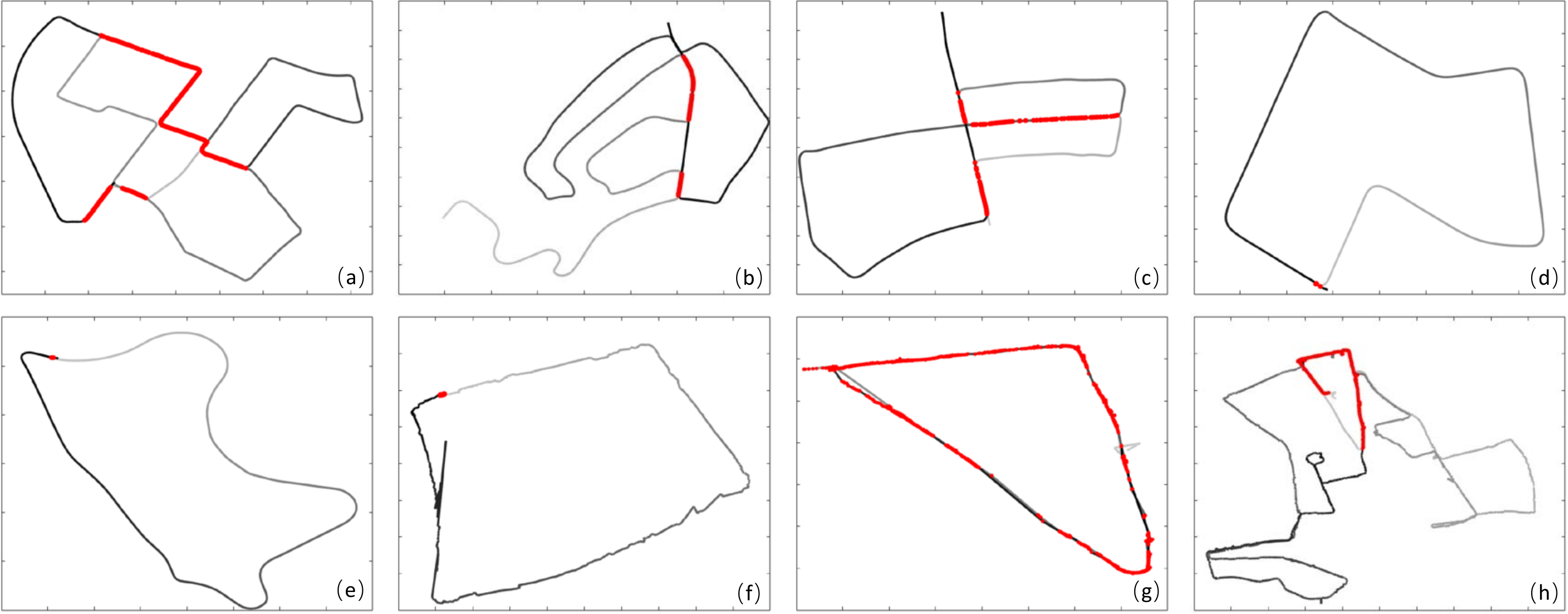}
	\end{center}
	\caption{Examples of place recognition on public dataset. The trajectory from GPS is plotted in from light grey to dark grey with time going. The loop closure reported by the proposed method is marked with red dot. (a, b, c, d, e) are collected from KITTI sequence 00, 02, 05, 07 and 09 respectively. (f, g, h) are collected from Oxford RobotCar sequence 2014\_05\_14\_13\_50\_20, 2014\_08\_11\_10\_22\_21 and 2014\_12\_02\_15\_30\_08 respectively. Our method achieves 80\% recall rate with no false positive.}
	\label{fig:KITTI Loop Closure Result}
\end{figure*}

\subsubsection{Training}
Given desired correlation output $\mathbf{g}$, we expect an optimal $\mathbf{h}$ that maps the kernel matrix $\boldsymbol{\kappa}_\mathbf{z}(\mathbf{x})$ to $\mathbf{g}$ such that the sum of squared error (SSE) is minimized in frequency domain:
 \begin{equation}
     \min\limits_{\hat{\mathbf{h}}^*}  
     \|\hat{\boldsymbol{\kappa}}_{\mathbf{z}}(\mathbf{x}) \varodot \hat{\mathbf{h}}^* - \hat{\mathbf{g}}\|^2  + \lambda\|\hat{\mathbf{h}}^*\|^2.
 \end{equation}
It has a closed-form solution:
\begin{equation}
    \hat{\mathbf{h}}^* = \frac{\hat{\mathbf{g}}}{ \hat{\boldsymbol{\kappa}}_{\mathbf{z}}(\mathbf{x}) + \lambda}, \label{eqn:optimal h}
\end{equation}
where operator $\frac{\cdot}{\cdot}$ is the element-wise division.
In the training stage, the optimal correlator $\mathbf{h}$ is solved with $\mathbf{x}=\mathbf{z}$. Note that due to the efficiency of \eqref{eqn:optimal h}, we can obtain $\hat{\mathbf{h}}^*$ online, hence no pre-trained database is required.

\subsubsection{Retrieval}
As shown in \cite{wang2018kernel}, kernel cross-correlation is equivariant to affine transforms, \textit{i.e.}, if input $\mathbf{x}$ is transformed (translation, scale, and rotation), the output $\mathbf{g}$ will be translated accordingly.
Therefore, the transform of a test sample $\mathbf{x}$ can be estimated by examining the translation of correlation output in \eqref{eq:kcc}.
For simplicity, we set $\mathbf{g}$ in \eqref{eqn:optimal h} as
\begin{equation}\label{eq:response}
    \mathbf{g}(x,y) = \left \{
    \begin{aligned}
        1, & ~\text{if}~ x=0, y=0\\
        0, & ~\text{otherwise}
    \end{aligned} \right.
\end{equation}
Because of noise and distortion, we cannot always obtain a pure translational response of $\mathbf{g}$ in \eqref{eq:response}.
Therefore, we can take the maximum value in the test output as the similarity $\zeta$ between training and test samples, which will be invariant to the affine transforms of $\mathbf{x}$. 
\begin{equation}
    \zeta = \max \mathcal{F}^{-1}\left( \hat{\boldsymbol{\kappa}}_\mathbf{z}(\mathbf{x})\varodot\hat{\mathbf{h}}^*\right). \label{eqn: kcc testing}
\end{equation}
The procedure of saliency retrieval using KCC is presented in \fref{fig: Loop Detected}, in which a loop closure candidate is detected.

\subsection{Consistency Verification}\label{sec:geometric}

As mentioned before, re-identification through saliency can improve the computational efficiency. However, the matched salient region may also cause false positive because the approach only identifies the similarity between query and candidate salient regions.
Therefore, it is necessary to check consistency before closing the loop since any false positive will lead to localization and mapping failure easily. 

The consistency verification is performed only on detected loop pairs by brute force matching using ORB features. Despite computationally expensive, matching on raw feature descriptor is able to remove most of the false positives produced by saliency re-identification, implying that traditional feature matching is a good complementary technique for saliency matching. We extract 1000 descriptors from current image and candidate image. A loop closure is determined when the number of matched descriptor pairs exceeds certain threshold.
Fortunately, the loop closure does not occur frequently and the computational cost is less significant in comparison to the entire process of place recognition. In the practical experiment, it costs less than 5\% of total computational expense so that such computational cost is worthy to guarantee the geometrical consistency.


\begin{table}[!b]
\begin{center}
\begin{tabular}{ccc}
\toprule
Parameter  & Description &  Value  \\ 
\midrule
$\sigma$   &  Standard deviation of Gaussian filter $\mathcal{G}_{\sigma}$ &  2.0\\
$n$   & Size of average filter  $\mathbf{f}_{ave}$   &  7       \\ 
$\phi_{th}$     &  Threshold of contrast density        &   58 \\
$\rho_{th}$     &  Threshold of edge complexity        &   0.5 \\
$\eta_{th}$   &  Threshold of KCC similarity test $\eta$ &  0.4 \\
\bottomrule
\end{tabular}
\caption{Parameter List.}
\label{table:parameter_list}
\end{center}
\end{table}
\section{Experiments}
\label{sec:experiment}

\subsection{Setup and Metric}
In our experiment, the proposed method and the compared methods are coded in C++ and tested on an \textit{Intel i7-8700 3.2 GHz} CPU. The source codes of compared methods are also included in provided link. The parameter list used in this paper is presented in Table \ref{table:parameter_list}. 

Three metrics, \textit{i.e.}, recall precision, recall rate, and computational cost, that are widely used to evaluate place recognition, are adopted in this paper.
Specifically, recall precision is the percentage of success match; recall rate is the reported matches against total matches obtained from ground truth; computational cost refers to the processing time. 
To note that, for practical applications, recall precision is required to be as high as possible since false positive will result in false mapping easily. Computational cost is of great concern as most of robotic systems do not possess enough computational power. Recall rate is considered as satisfied if most of the loop places are identified.
\begin{table}[!b]
    \begin{center}
    \begin{tabular}{ccc}
    \toprule
    Dataset & Description & Image Resolution  \\
    \midrule
    KITTI  \cite{geiger2013vision}  & Outdoor, dynamic &   370$\times$1226    \\ 
    Freiburg3 \cite{sturm12iros}   & Indoor, static &   1280$\times$1024   \\  
    New College \cite{CumminsNewmanIJRR08} & Outdoor, dynamic    &   640$\times$480 \\  
    City Center \cite{CumminsNewmanIJRR08} & Outdoor, dynamic    &   640$\times$480 \\  
    RobotCar \cite{maddern20171} & Outdoor, dynamic &   1280$\times$960 \\
    \bottomrule
    \end{tabular}
    \caption{Details of Datasets.}
    \label{table:datasets information.}
    \end{center}
\end{table}

\begin{figure*}[!t]
	\centering
	\begin{subfigure}[t]{0.32\textwidth}
		\centering
		\includegraphics[width=1.0\linewidth]{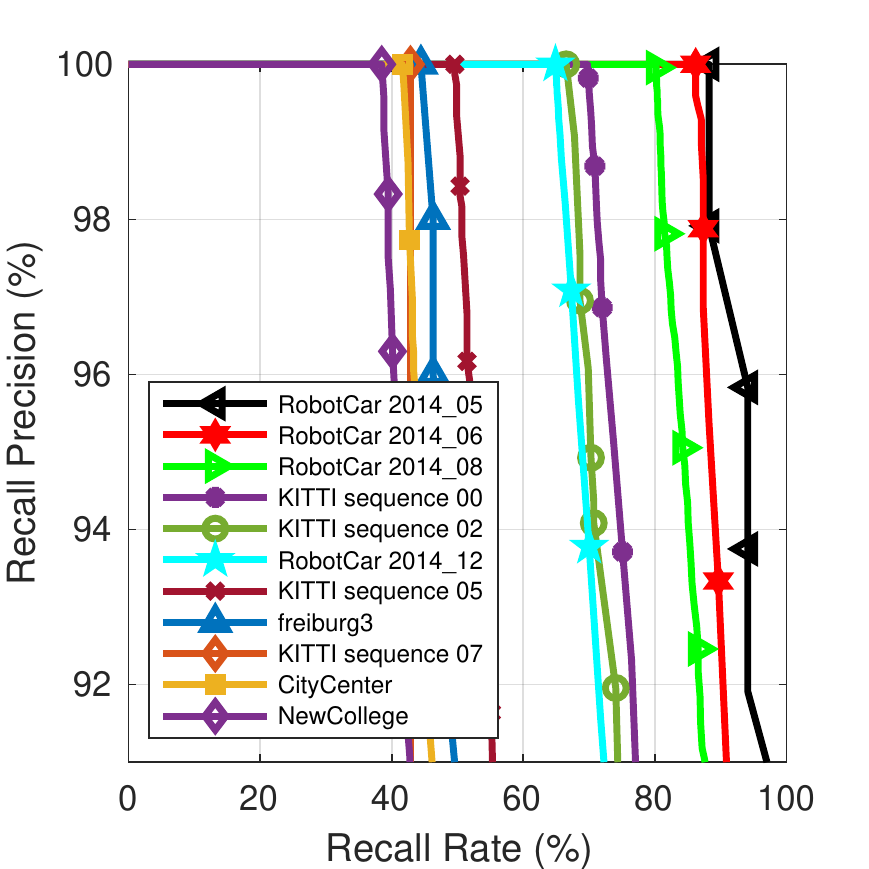}
		\caption{Precision-recall Curve.}
		\label{fig:precision_recall}
	\end{subfigure}
	\begin{subfigure}[t]{0.32\textwidth}
		\centering
		\includegraphics[width=1.0\linewidth]{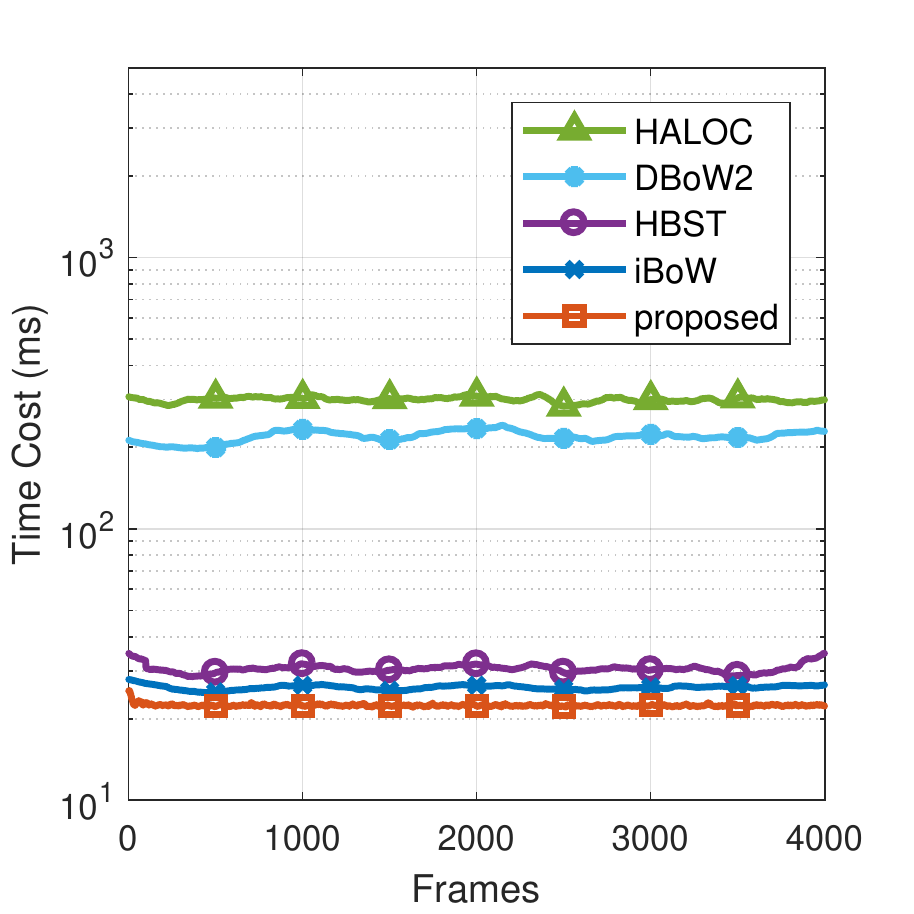}
		\caption{Feature/Saliency Extraction.}
		\label{fig:feature_extraction_comparison}
	\end{subfigure}%
	\begin{subfigure}[t]{0.32\textwidth}
		\centering
		\includegraphics[width=1.0\linewidth]{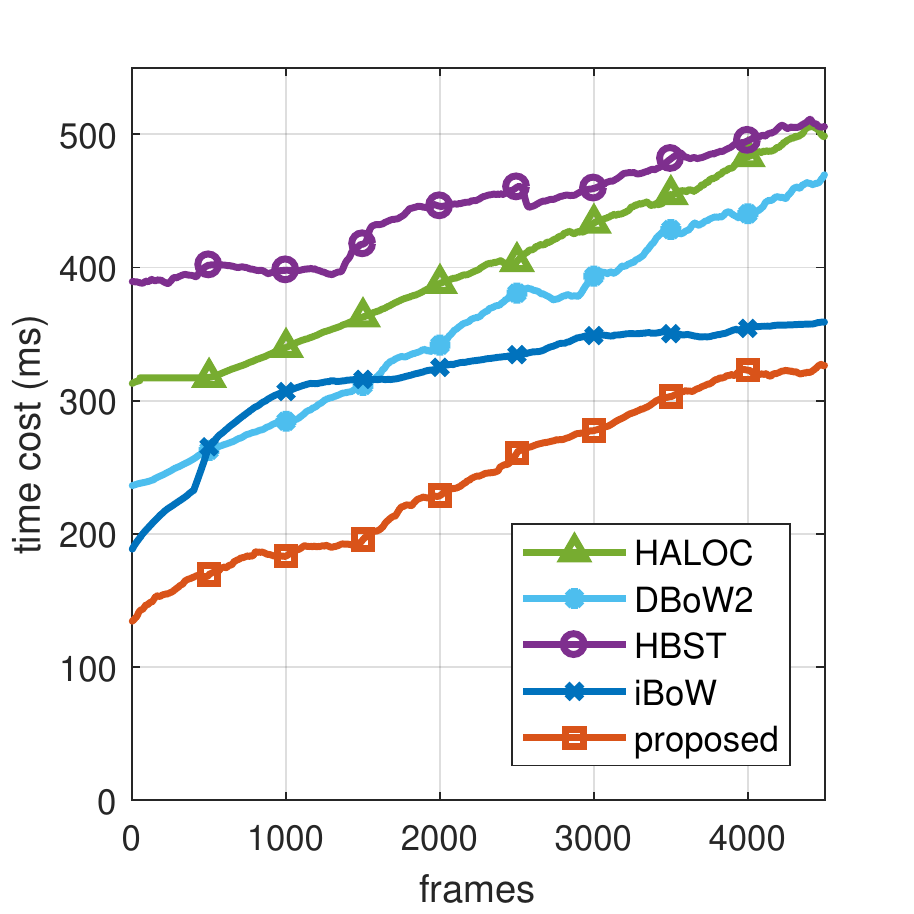}
		\caption{Running Time.}
		\label{fig:computation-cose}
	\end{subfigure}
	\caption{(a) Precision-recall curve of the proposed method on different dataset. (b) Comparison of saliency extraction with feature extraction methods in KITTI sequence 00. (c) Time cost comparison of different approaches on KITTI sequence 00; Our approach achieves much faster speed than existing methods.}
\end{figure*}

\subsection{Performance evaluation}
The evaluation is performed based on various public datasets that are popularly used for loop closure detection, including KITTI \cite{geiger2013vision}, TUM \cite{sturm12iros}, New College \cite{CumminsNewmanIJRR08}, City Center \cite{CumminsNewmanIJRR08}, and Oxford RobotCar \cite{maddern20171}. They include various scenarios consisting of indoor localization and autonomous driving, \textit{etc}. The datasets contain large scale collections of up to 35,000 frames (RobotCar dataset), medium scale collections of 5,000 frames (KITTI dataset) and small scale collections of 1,000 frames (New College dataset). The details are listed in \tref{table:datasets information.}. The ground truth of loop closure detection is collected based on GPS/Vicon information.

\paragraph{Results}
\fref{fig:KITTI Loop Closure Result} shows the examples of loop closure detection by the proposed approach. We plot the results from large datasets for illustration. The datasets include more than 60,000 images in total and the largest dataset contains around 35,000 images. 
As can be seen, most of re-visited places are identified and there is no false positive. 

\paragraph{Precision-Recall Curve}
The precision-recall curve on all datasets is shown in \fref{fig:precision_recall}. In particular, RobotCar \seqsplit{2014\_05}, \seqsplit{2014\_06}, \seqsplit{2014\_08}, and 2014\_12 refers to RobotCar sequence \seqsplit{2014\_05\_14\_13\_50\_20}, \seqsplit{2014\_06\_26\_09\_53\_12}, \seqsplit{2014\_08\_11\_10\_22\_21}, and \seqsplit{2014\_12\_02\_15\_30\_08}, respectively. It is notable that our method achieves 100\% precision over all datasets. In some challenging outdoor dynamic environments, \textit{e.g.}, Oxford RobotCar dataset with 35,000 images, our approach achieves high recall rate up to 90\%. More importantly, our approach is able to run up to 50\hertz, which is computationally cost-effective for many robotic applications.




\subsection{Comparison}
\begin{figure}[!b]
	\begin{center}
		\includegraphics[width=0.85\linewidth]{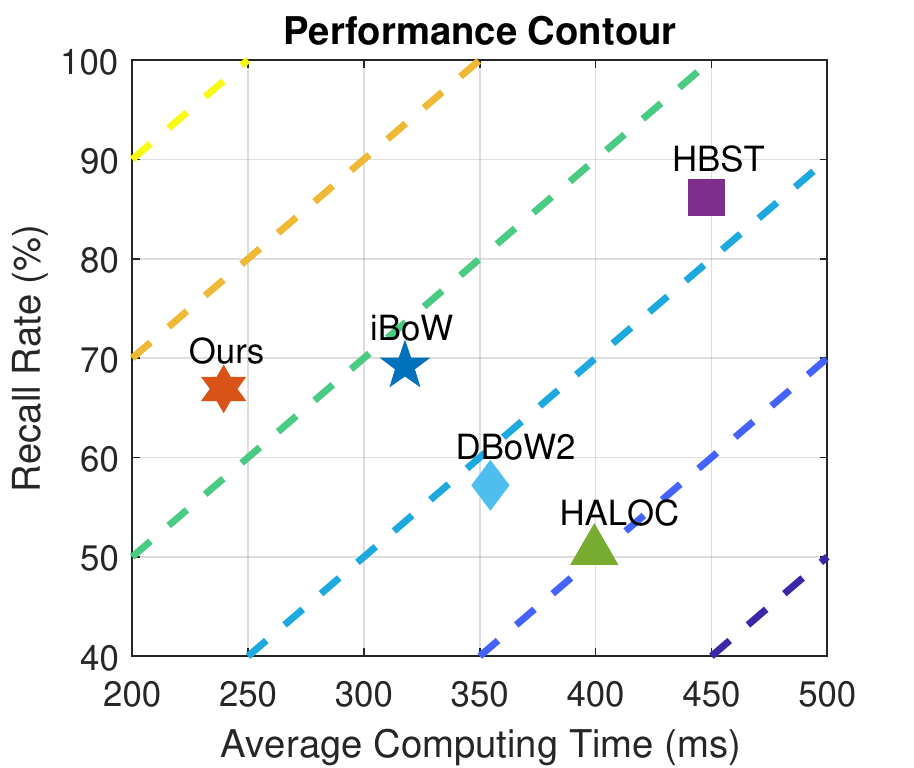}
	\end{center}
	\caption{Overall performance of different approaches.}
	\label{fig:metrics_comparison}
\end{figure}
To further illustrate the efficiency of the proposed method, we compare the state-of-the-art feature-based methods such as iBoW \cite{garcia2018ibow}, HALOC \cite{carrasco2016global}, DBoW2 \cite{galvez2012bags}, and HBST \cite{schlegel2018hbst}. 
In particular, DBoW2 appears in the early stage and yet is still one of the most popular methods and has been used for ORB SLAM \cite{mur2017orb} and LDSO \cite{gao2018ldso}. The rest methods are introduced in the recent years and they have been shown to be effective in loop closure detection. For all compared methods, one thousand local features, which is a typical recommended feature extraction number, are extracted for each frame. To note that, HBST is evaluated at quarter of original video size due to its huge memory consumption and time cost.

\paragraph{Efficiency Evaluation}
For computational efficiency, we use the KITTI sequence 00 for example since the speed comparison is similar across different datasets. \fref{fig:feature_extraction_comparison} shows the performance of the saliency extraction compared to feature extraction from other methods. 
As mentioned before, saliency extraction is only of linear computational complexity in frequency domain, it is much faster than local feature extraction. In iBoW and HBST, raw ORB feature is used and hence the computational cost is also low. However, HALOC and DBoW2 require further dimension reduction hence the feature extraction time is higher.
\fref{fig:computation-cose} shows the overall computational cost of respective methods. Our approach achieves the fastest speed among all the approaches, \textit{i.e.}, a few times faster than DBoW2 and HALOC.

\paragraph{Overall Performance}
\fref{fig:metrics_comparison} compares the overall performance of different methods in KITTI sequence 00. HBST that performs feature match on raw image descriptors achieves the highest recall rate.
However, both the computational cost and memory cost are too high to build the binary search tree so that it can be difficult to be implemented in practical application. 
In comparison, our approach provides a feasible solution for visual place recognition that is computationally efficient and accurate enough.
Therefore, our approach achieves a good trade-off between both precision and efficiency.
It is also notable that our method requires no offline training or vocabulary loading, making it more feasible for robotic real-time applications.

\section{Conclusion}
\label{sec:conclusion}

In this paper, we present a novel framework for visual place recognition based on saliency re-identification. 
It mainly consists of two components, \textit{i.e.} saliency detection and retrieval.
To reduce the computational cost, both tasks are performed in frequency domain to take advantage of the efficiency of element-wise operation, resulting in an overall computational complexity of $\mathcal{O}(N\log N)$. 
The proposed method is open sourced as a C++ library.
The experiments show that our method is more computationally efficient and accurate compared with the other state-of-the-art methods.\underline{}


\balance
\bibliographystyle{IEEEtran}
\bibliography{IEEEabrv,references}

\begin{thebibliography}{10}
\providecommand{\url}[1]{#1}
\csname url@rmstyle\endcsname
\providecommand{\newblock}{\relax}
\providecommand{\bibinfo}[2]{#2}
\providecommand\BIBentrySTDinterwordspacing{\spaceskip=0pt\relax}
\providecommand\BIBentryALTinterwordstretchfactor{4}
\providecommand\BIBentryALTinterwordspacing{\spaceskip=\fontdimen2\font plus
\BIBentryALTinterwordstretchfactor\fontdimen3\font minus
  \fontdimen4\font\relax}
\providecommand\BIBforeignlanguage[2]{{%
\expandafter\ifx\csname l@#1\endcsname\relax
\typeout{** WARNING: IEEEtran.bst: No hyphenation pattern has been}%
\typeout{** loaded for the language `#1'. Using the pattern for}%
\typeout{** the default language instead.}%
\else
\language=\csname l@#1\endcsname
\fi
#2}}

\bibitem{lowry2015visual}
S.~Lowry, N.~S{\"u}nderhauf, P.~Newman, J.~J. Leonard, D.~Cox, P.~Corke, and
  M.~J. Milford, ``Visual place recognition: A survey,'' \emph{IEEE
  Transactions on Robotics}, vol.~32, no.~1, pp. 1--19, 2015.

\bibitem{wang2020intensity}
H.~Wang, C.~Wang, and L.~Xie, ``Intensity scan context: Coding intensity and
  geometry relations for loop closure detection,'' in \emph{IEEE international
  conference on robotics and automation (ICRA)}, 2020.

\bibitem{nguyen2020loosely}
T.~H. Nguyen, T.-M. Nguyen, M.~Cao, and L.~Xie, ``Loosely-coupled
  ultra-wideband-aided scale correction for monocular visual odometry,''
  \emph{Unmanned Systems}, vol.~8, no.~02, pp. 179--190, 2020.

\bibitem{lowe2004distinctive}
D.~G. Lowe, ``Distinctive image features from scale-invariant keypoints,''
  \emph{International journal of computer vision}, vol.~60, no.~2, pp. 91--110,
  2004.

\bibitem{arroyo2015towards}
R.~Arroyo, P.~F. Alcantarilla, L.~M. Bergasa, and E.~Romera, ``Towards
  life-long visual localization using an efficient matching of binary sequences
  from images,'' in \emph{2015 IEEE international conference on robotics and
  automation (ICRA)}.\hskip 1em plus 0.5em minus 0.4em\relax IEEE, 2015, pp.
  6328--6335.

\bibitem{cieslewski2017efficient}
T.~Cieslewski and D.~Scaramuzza, ``Efficient decentralized visual place
  recognition using a distributed inverted index,'' \emph{IEEE Robotics and
  Automation Letters}, vol.~2, no.~2, pp. 640--647, 2017.

\bibitem{rublee2011orb}
E.~Rublee, V.~Rabaud, K.~Konolige, and G.~Bradski, ``Orb: An efficient
  alternative to sift or surf,'' in \emph{Computer Vision (ICCV), 2011 IEEE
  international conference on}.\hskip 1em plus 0.5em minus 0.4em\relax IEEE,
  2011, pp. 2564--2571.

\bibitem{leutenegger2011brisk}
S.~Leutenegger, M.~Chli, and R.~Siegwart, ``Brisk: Binary robust invariant
  scalable keypoints,'' in \emph{2011 IEEE international conference on computer
  vision (ICCV)}.\hskip 1em plus 0.5em minus 0.4em\relax Ieee, 2011, pp.
  2548--2555.

\bibitem{schlegel2018hbst}
D.~Schlegel and G.~Grisetti, ``Hbst: A hamming distance embedding binary search
  tree for visual place recognition,'' \emph{arXiv preprint arXiv:1802.09261},
  2018.

\bibitem{garcia2018ibow}
E.~Garcia-Fidalgo and A.~Ortiz, ``ibow-lcd: An appearance-based loop-closure
  detection approach using incremental bags of binary words,'' \emph{IEEE
  Robotics and Automation Letters}, vol.~3, no.~4, pp. 3051--3057, 2018.

\bibitem{cummins2008fab}
M.~Cummins and P.~Newman, ``Fab-map: Probabilistic localization and mapping in
  the space of appearance,'' \emph{The International Journal of Robotics
  Research}, vol.~27, no.~6, pp. 647--665, 2008.

\bibitem{geiger2013vision}
A.~Geiger, P.~Lenz, C.~Stiller, and R.~Urtasun, ``Vision meets robotics: The
  kitti dataset,'' \emph{The International Journal of Robotics Research},
  vol.~32, no.~11, pp. 1231--1237, 2013.

\bibitem{cong2018review}
R.~Cong, J.~Lei, H.~Fu, M.-M. Cheng, W.~Lin, and Q.~Huang, ``Review of visual
  saliency detection with comprehensive information,'' \emph{IEEE Transactions
  on Circuits and Systems for Video Technology}, 2018.

\bibitem{donk2008effects}
M.~Donk and W.~van Zoest, ``Effects of salience are short-lived,''
  \emph{Psychological Science}, vol.~19, no.~7, pp. 733--739, 2008.

\bibitem{wang2020visual}
C.~Wang, W.~Wang, Y.~Qiu, Y.~Hu, and S.~Scherer, ``{Visual Memorability for
  Robotic Interestingness via Unsupervised Online Learning},'' in
  \emph{European Conference on Computer Vision (ECCV)}, 2020.

\bibitem{hou2007saliency}
X.~Hou and L.~Zhang, ``Saliency detection: A spectral residual approach,'' in
  \emph{Computer Vision and Pattern Recognition, 2007. CVPR'07. IEEE Conference
  on}.\hskip 1em plus 0.5em minus 0.4em\relax IEEE, 2007, pp. 1--8.

\bibitem{glover2012openfabmap}
A.~Glover, W.~Maddern, M.~Warren, S.~Reid, M.~Milford, and G.~Wyeth,
  ``Openfabmap: An open source toolbox for appearance-based loop closure
  detection,'' in \emph{Robotics and automation (ICRA), 2012 IEEE international
  conference on}.\hskip 1em plus 0.5em minus 0.4em\relax IEEE, 2012, pp.
  4730--4735.

\bibitem{bay2006surf}
H.~Bay, T.~Tuytelaars, and L.~Van~Gool, ``Surf: Speeded up robust features,''
  in \emph{European conference on computer vision}.\hskip 1em plus 0.5em minus
  0.4em\relax Springer, 2006, pp. 404--417.

\bibitem{galvez2012bags}
D.~G{\'a}lvez-L{\'o}pez and J.~D. Tardos, ``Bags of binary words for fast place
  recognition in image sequences,'' \emph{IEEE Transactions on Robotics},
  vol.~28, no.~5, pp. 1188--1197, 2012.

\bibitem{carrasco2016global}
P.~L.~N. Carrasco, F.~Bonin-Font, and G.~Oliver-Codina, ``Global image
  signature for visual loop-closure detection,'' \emph{Autonomous Robots},
  vol.~40, no.~8, pp. 1403--1417, 2016.

\bibitem{gehrig2017visual}
M.~Gehrig, E.~Stumm, T.~Hinzmann, and R.~Siegwart, ``Visual place recognition
  with probabilistic voting,'' in \emph{2017 IEEE International Conference on
  Robotics and Automation (ICRA)}.\hskip 1em plus 0.5em minus 0.4em\relax IEEE,
  2017, pp. 3192--3199.

\bibitem{wang2019kervolutional}
C.~Wang, J.~Yang, L.~Xie, and J.~Yuan, ``Kervolutional neural networks,'' in
  \emph{The IEEE Conference on Computer Vision and Pattern Recognition (CVPR)},
  2019, pp. 31--40.

\bibitem{chen2017deep}
Z.~Chen, A.~Jacobson, N.~S{\"u}nderhauf, B.~Upcroft, L.~Liu, C.~Shen, I.~Reid,
  and M.~Milford, ``Deep learning features at scale for visual place
  recognition,'' in \emph{2017 IEEE International Conference on Robotics and
  Automation (ICRA)}.\hskip 1em plus 0.5em minus 0.4em\relax IEEE, 2017, pp.
  3223--3230.

\bibitem{arandjelovic2016netvlad}
R.~Arandjelovic, P.~Gronat, A.~Torii, T.~Pajdla, and J.~Sivic, ``Netvlad: Cnn
  architecture for weakly supervised place recognition,'' in \emph{Proceedings
  of the IEEE conference on computer vision and pattern recognition}, 2016, pp.
  5297--5307.

\bibitem{hausler2019multi}
S.~Hausler, A.~Jacobson, and M.~Milford, ``Multi-process fusion: Visual place
  recognition using multiple image processing methods,'' \emph{IEEE Robotics
  and Automation Letters}, vol.~4, no.~2, pp. 1924--1931, 2019.

\bibitem{dalal2005histograms}
N.~Dalal and B.~Triggs, ``Histograms of oriented gradients for human
  detection,'' in \emph{Proceedings of the IEEE Computer Society Conference on
  Computer Vision and Pattern Recognition (CVPR)}, 2005.

\bibitem{srivastava2003advances}
A.~Srivastava, A.~B. Lee, E.~P. Simoncelli, and S.-C. Zhu, ``On advances in
  statistical modeling of natural images,'' \emph{Journal of mathematical
  imaging and vision}, vol.~18, no.~1, pp. 17--33, 2003.

\bibitem{reddy1996fft}
B.~S. Reddy and B.~N. Chatterji, ``An fft-based technique for translation,
  rotation, and scale-invariant image registration,'' \emph{IEEE transactions
  on image processing}, vol.~5, no.~8, pp. 1266--1271, 1996.

\bibitem{wang2018kernel}
C.~Wang, L.~Zhang, L.~Xie, and J.~Yuan, ``Kernel cross-correlator,'' in
  \emph{Thirty-Second AAAI Conference on Artificial Intelligence (AAAI)}, 2018.

\bibitem{wang2018correlation}
C.~Wang, T.~Ji, T.-M. Nguyen, and L.~Xie, ``Correlation flow: robust optical
  flow using kernel cross-correlators,'' in \emph{2018 IEEE International
  Conference on Robotics and Automation (ICRA)}.\hskip 1em plus 0.5em minus
  0.4em\relax IEEE, 2018, pp. 836--841.

\bibitem{sturm12iros}
J.~Sturm, N.~Engelhard, F.~Endres, W.~Burgard, and D.~Cremers, ``A benchmark
  for the evaluation of rgb-d slam systems,'' in \emph{Proc. of the
  International Conference on Intelligent Robot Systems (IROS)}, Oct. 2012.

\bibitem{CumminsNewmanIJRR08}
M.~Cummins and P.~Newman, ``{FAB-MAP: Probabilistic Localization and Mapping in
  the Space of Appearance},'' \emph{The International Journal of Robotics
  Research}, vol.~27, no.~6, pp. 647--665, 2008.

\bibitem{maddern20171}
W.~Maddern, G.~Pascoe, C.~Linegar, and P.~Newman, ``1 year, 1000 km: The oxford
  robotcar dataset,'' \emph{The International Journal of Robotics Research},
  vol.~36, no.~1, pp. 3--15, 2017.

\bibitem{mur2017orb}
R.~Mur-Artal and J.~D. Tard{\'o}s, ``Orb-slam2: An open-source slam system for
  monocular, stereo, and rgb-d cameras,'' \emph{IEEE Transactions on Robotics},
  vol.~33, no.~5, pp. 1255--1262, 2017.

\bibitem{gao2018ldso}
X.~Gao, R.~Wang, N.~Demmel, and D.~Cremers, ``Ldso: Direct sparse odometry with
  loop closure,'' in \emph{2018 IEEE/RSJ International Conference on
  Intelligent Robots and Systems (IROS)}.\hskip 1em plus 0.5em minus
  0.4em\relax IEEE, 2018, pp. 2198--2204.

\end{thebibliography}

\end{document}